\documentclass[11pt,a4paper]{article}

\usepackage[utf8]{inputenc}
\usepackage[T1]{fontenc}
\usepackage{graphicx}
\usepackage{amsmath,amssymb,amsfonts}
\usepackage{booktabs}
\usepackage{hyperref}
\usepackage{geometry}
\usepackage{authblk}
\usepackage{natbib}
\usepackage{abstract}

\title{\textbf{Style2Code: A Dual-Modal Contrastive Learning Framework for Style-Controllable Code Generation}}

\author[1]{Dutao Zhang}
\author[2]{YuLong He}
\author[3]{Nicolas Rafael Arroyo Arias}
\author[4]{Boyi Li}
\author[1]{ Kovalchuk\thanks{Corresponding author: \texttt{sergey.v.kovalchuk@gmail.com}}}
\affil[1]{ITMO University, Saint Petersburg, Russia}
\affil[2]{Saint Petersburg University, Saint Petersburg, Russia}

\begin{document}

\maketitle

\begin{abstract}
Controllable code generation aims to produce source code that adheres to specified style patterns while maintaining functional correctness. This research direction primarily serves scenarios where programmers use code generation models in daily practice, enabling models to better adapt to users' specific project requirements. Additionally, controllable code generation allows models to produce code with more consistent style that aligns closely with developers' coding habits, thereby achieving alignment between model outputs and user coding practices. However, existing methods often rely on multi-model collaborative training, requiring multiple AI systems to mutually correct errors to optimize outputs, which not only increases training and deployment complexity but also limits flexibility and scalability in practical applications. To address this, we propose Style2Code, a novel dual-modal framework for style-controllable code generation. Our approach encodes code style as an explicit 34-dimensional vector representation that captures fine-grained style attributes from three dimensions: naming conventions (14-dim), code spacing layout (9-dim), and structural complexity (11-dim), including aspects such as naming conventions, indentation patterns, and structural layout. Subsequently, this style vector is concatenated with source code tokens and fed into a pre-trained language model (e.g., Flan-T5), enabling the decoder to generate code reflecting the target style. Experimental results on multiple benchmark datasets demonstrate that Style2Code achieves significant improvements in both style alignment and generation quality, outperforming state-of-the-art baseline methods on BLEU, ROUGE, and Code Style Similarity (CSS) metrics. To our knowledge, Style2Code is the first model to achieve explicit and user-controllable code style transfer through continuous vector conditioning in a dual-modal generation framework.
\end{abstract}


\noindent\textbf{Source code and dataset:} \url{https://github.com/zh19980811/Style2Code}

\section{Introduction}
\label{sec:intro}

\subsection{Problem and Motivation}

Developers frequently observe that AI-generated code may not fully meet their requirements in terms of style when using AI models for daily code generation. This style divergence primarily stems from the diversity of training data and differences in user prompts. Furthermore, different models may generate code with slightly different styles. When users do not provide code examples, specify style requirements in prompts, or give detailed algorithmic requirements, model outputs often exhibit significant randomness \citep{zou2019code}.

In practical software development, inconsistencies between AI-generated code style and established project conventions can lead to a series of negative consequences. First, inconsistent naming conventions force developers to memorize additional variants, thereby increasing cognitive load and reducing team collaboration efficiency \citep{chad2025ai}. Second, differences in indentation and formatting divert attention from logical correctness during code review, requiring additional effort for style correction and frequently introducing conflicts during pull request integration \citep{hindle2008indentation}. Finally, AI-specific code smells—characteristic patterns in AI-generated code that may indicate underlying quality issues, such as irreproducibility, silent failures, or poor model generalization—are easily overlooked when using AI-generated code \citep{mahmoudi2025ai}. By allowing users to guide AI code generation with specific conditions, the generated code can better conform to project requirements and coding standards, thereby reducing the workload required for subsequent code maintenance.

\subsection{Motivation and Significance}

Our primary goal is to enable AI models to generate code that aligns with users' personal or project-specific coding styles during the generation process. The key component in achieving this capability is code style—a consistent and recognizable pattern that defines how code is written and structured. These patterns include indentation schemes, naming conventions, common structural layouts, and preferred utility functions, all of which reflect developers' habitual practices. By following these patterns, AI-generated code becomes more readable, maintainable, and seamlessly integrates into existing codebases.

In professional development environments, this style uniformity is not merely an aesthetic requirement but a functional necessity—it improves collaboration, reduces code review costs, and maintains overall consistency in large software projects. Established conventions, such as Google Style Guide, Python's PEP 8, or Airbnb's JavaScript Style Guide, demonstrate that style specifications help all contributors work under the same set of assumptions. Inspired by these principles, our approach aims to encode and reproduce such style regularities through style-aware representation learning, ensuring that generated code is not only functionally correct but also conforms to user-established style specifications. This alignment bridges the gap between machine-generated and human-written code, making AI-assisted programming more natural, efficient, and trustworthy.

\subsection{Contributions}

Our work applies cross-modal contrastive learning to the code style transfer task. In evaluation on the same task, we outperform the state-of-the-art model MPCoder \citep{dai2024mpcoder} on multiple metrics. Through contrastive learning, we achieve mutual recognition between code and its style encoding, addressing the cold-start problem. Additionally, we solve the challenge of clustering diverse code samples. Our main contributions are summarized as follows:

\begin{enumerate}
\item We design a fine-grained explicit style encoder that directly extracts interpretable style features from raw source code—such as indentation schemes, whitespace alignment, loop nesting depth, and structural repetition patterns—without relying on manual annotations or handcrafted templates.

\item We introduce a contrastive alignment framework that semantically aligns code embeddings with their corresponding style vectors, enabling the model to generalize to unseen styles while maintaining controllability during generation. This is similar in spirit to CLIP \citep{radford2021learning}, but actually combines style learning and code generation tasks.

\item We propose a two-stage training paradigm that first independently trains the style encoder to obtain stable and accurate style representations, then injects these representations into the code generator for style-controlled generation. This decoupled training strategy improves overall training stability and significantly enhances the effectiveness of style transfer.

\item Furthermore, we formulate the style transfer task as a dual-modal learning problem and explore its potential in the context of personalized code generation, achieving flexible and user-controllable style variation.
\end{enumerate}

\section{Related Work}
\label{sec:related}

\subsection{Personalized Code Generation}

Personalized code generation has emerged as an important research direction in automated code synthesis. MPCoder \citep{dai2024mpcoder} represents a significant advance, integrating both explicit and implicit style representations for multi-user scenarios. However, it faces several limitations: (1) difficulty in generalizing implicit style cues embedded in code semantics; (2) reliance on fixed handcrafted style encodings and metrics, which limits adaptability; (3) requirement for large user-specific datasets, leading to cold-start and scalability issues. These constraints hinder its deployment in data-limited or constantly evolving dynamic environments. In contrast, our framework aims to achieve style control through transferable representation learning without requiring user-specific retraining.

\subsection{Code Style Representation and Authorship Attribution}

Several studies have attempted to model style information in code through language-based or author-aware learning. Ahmed et al. \citep{ahmed2024automatic} enhance LLM style representations by integrating semantic information through code summarization. Similarly, Choi and Tan \citep{choi2025find} explore source code authorship attribution using large language models, highlighting applications in software forensics and plagiarism detection. Their findings demonstrate that LLMs can capture consistent style patterns across different programming languages, achieving competitive author classification performance under zero-shot and few-shot prompting. Although these works focus on author identification rather than code generation, they indicate that modern LLMs inherently encode style features that can be leveraged for personalized generation.

\subsection{Contrastive Learning and Style-Aware Learning}

Recent advances in contrastive learning have further inspired our approach. Bui et al. \citep{bui2021self} propose a semi-supervised contrastive framework that trains source code encoders to distinguish semantically similar and dissimilar code snippets, enhancing their understanding of structural and functional relationships. Building on this, Liu and Nguyen et al. \citep{liu2024learning} employ iterative synthesis and questioning methods to refine code style representations, improving model adaptability and style consistency. Unlike these methods, our proposed Style2Code framework performs dual-modal contrastive learning between style and code embeddings, achieving controllable style alignment without requiring large amounts of labeled data.

Based on these insights, the next section introduces the overall architecture and training objectives of Style2Code, which integrates style representation learning and code generation in a unified framework.

\section{Method}
\label{sec:method}

We define the personalized code generation task as a dual-modal generation problem, where the goal is to generate code that preserves the original functionality while adhering to user-specified coding styles. One modality provides functional semantics, while the other provides style guidance \citep{huang2024accelerating}.

\subsection{Task Definition}

Given a source code snippet and a reference example reflecting the target style (while maintaining equivalent functionality), the generation process involves two modalities: (1) the textual representation of the reference code, and (2) a quantitative style vector extracted by a dedicated style encoding module. During training, the model learns to align these modalities so that it can later generate a version that is functionally equivalent to the source code but style-consistent, reflecting the characteristics of the reference. This formulation enables fine-grained and continuous control over various aspects of code style \citep{han2023text}—such as formatting, indentation, naming conventions, and structural organization—without compromising semantic correctness.

\subsection{Proposed Framework}

To overcome the limitations of existing methods, we propose a novel two-stage framework for style-controllable code generation. This framework explicitly extracts style representations from target style reference code and uses these representations to guide a fine-tuned language model in generating functionally equivalent but style-aligned code.

Given a source code snippet and a reference code sample (with similar functionality but desired style), we first use a style analyzer to extract a 34-dimensional style vector from the reference code and encode it through a style encoder. The resulting embedding captures latent style features such as naming conventions, indentation patterns, whitespace density, and layout organization. This encoded vector is fused with the source code input and passed to a style-aware language model for generation \citep{toshevska2022review}.

The Style2Code framework comprises two main components: a style encoder (approximately 1.65 million parameters) and a style-controlled generator (based on Flan-T5-Large, approximately 850 million parameters). Overall, the complete model contains approximately 852 million parameters.

\begin{figure}[ht]
\centering
\includegraphics[width=0.9\textwidth]{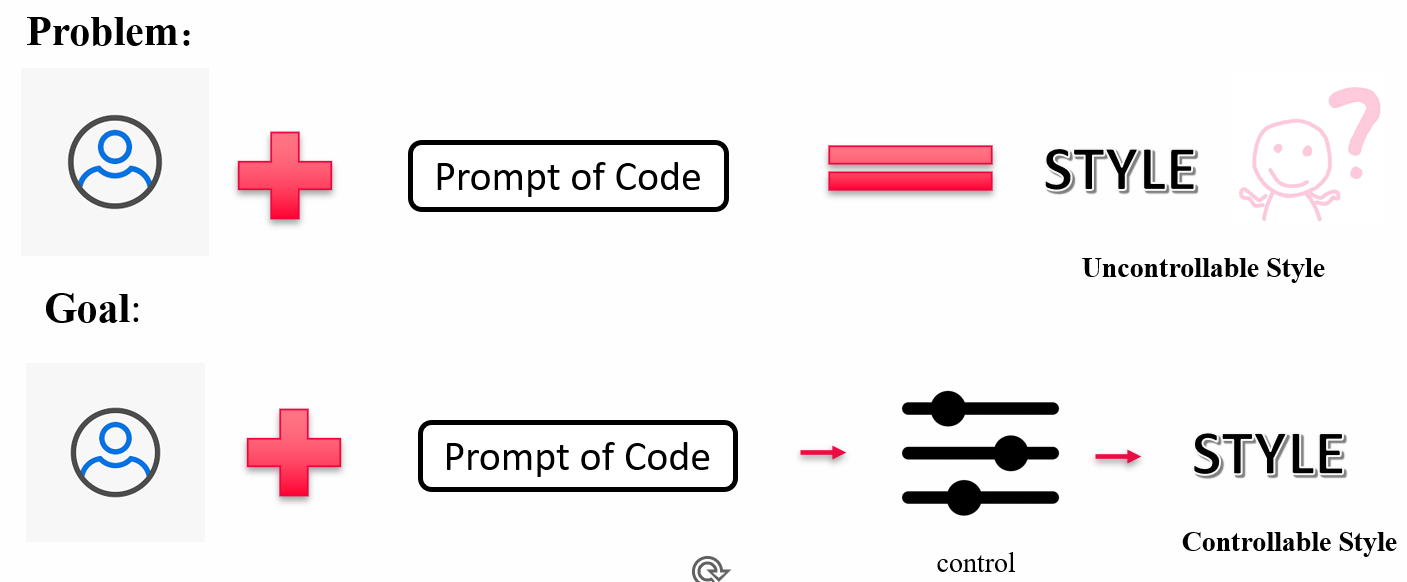}
\caption{Overall architecture of the Style2Code framework. The framework consists of two main components: (1) Style Encoder that extracts and encodes 34-dimensional style vectors from reference code, and (2) Style-Controlled Generator that takes source code and style embeddings to produce style-aligned output code.}
\label{fig:framework}
\end{figure}

\subsection{Style Analyzer}

The style analyzer is designed to explicitly extract structured and interpretable features that describe developers' habitual coding patterns. It organizes style modeling into three dimensions: naming style, spacing and layout style, and structural/functional style. These extracted features form a 34-dimensional vector $\mathbf{s} \in \mathbb{R}^{34}$, which serves as input to the style encoder.

\subsubsection{Naming Style}

Table~\ref{tab:naming} summarizes the naming-related features used for style encoding. These metrics capture statistical and categorical naming preferences that reflect individual developer or project styles.

\begin{table}[ht]
\caption{Naming Style Analysis Features}
\label{tab:naming}
\centering
\small
\begin{tabular}{@{}p{0.35\textwidth}p{0.55\textwidth}@{}}
\toprule
\textbf{Feature} & \textbf{Description} \\
\midrule
name\_length & Average name length \\
is\_snake\_case & Whether identifiers use snake\_case \\
style\_stat\_underscore\_ratio & Underscore usage ratio \\
style\_stat\_digit\_ratio & Digit ratio in identifiers \\
style\_stat\_symbol\_ratio & Special character ratio \\
style\_stat\_uppercase\_ratio & Uppercase letter ratio \\
style\_stat\_lowercase\_ratio & Lowercase letter ratio \\
style\_dist\_PascalCase & PascalCase name frequency \\
style\_dist\_snake\_case & snake\_case name frequency \\
style\_dist\_camelCase & camelCase name frequency \\
style\_dist\_UPPER\_CASE & UPPER\_CASE name frequency \\
style\_dist\_private & Private variable prefix (\_) usage \\
style\_dist\_dunder\_method & Dunder method (\_\_init\_\_) usage \\
naming\_consistency & Naming style consistency score \\
\bottomrule
\end{tabular}
\end{table}

\subsubsection{Structural and Functional Style}

This set of features is extracted through static analysis based on Abstract Syntax Trees (AST), quantifying code complexity and structural organization \citep{rabinovich2017abstract}. These metrics capture developers' function-level and control-flow preferences (Table~\ref{tab:structure}).

\begin{table}[ht]
\caption{Structural and Functional Style Features}
\label{tab:structure}
\centering
\small
\begin{tabular}{@{}p{0.35\textwidth}p{0.55\textwidth}@{}}
\toprule
\textbf{Feature} & \textbf{Description} \\
\midrule
call\_depth & Maximum function call depth \\
branch\_count & Number of branches \\
return\_count & Number of return statements \\
arg\_count & Number of function arguments \\
length & Lines of code per function \\
has\_docstring & Presence of docstring \\
has\_try\_except & Presence of try/except block \\
exception\_score & Exception handling specificity \\
redundancy\_ratio & Redundant code ratio \\
annotation\_ratio & Ratio of parameters with type hints \\
control\_structures & Number of control structures \\
\bottomrule
\end{tabular}
\end{table}

\subsubsection{Spacing and Layout Style}

Layout style focuses on whitespace, indentation, and comment usage. We define a visual descriptor called code space pattern, which transforms code into a fixed-size binary matrix representing indentation and spacing regularities. This captures developer layout habits that contribute to visual readability (Table~\ref{tab:layout}).

\begin{table}[ht]
\caption{Spacing and Layout Style Features}
\label{tab:layout}
\centering
\small
\begin{tabular}{@{}p{0.35\textwidth}p{0.55\textwidth}@{}}
\toprule
\textbf{Feature} & \textbf{Description} \\
\midrule
blank\_line\_count & Number of blank lines \\
line\_length\_avg & Average line length \\
line\_length\_variance & Line length variance \\
indentation\_level\_avg & Average indentation level \\
space\_before\_operator & Space usage before operators \\
comment\_ratio & Ratio of comment lines to total lines \\
type\_hint\_ratio & Ratio of parameters with explicit type hints \\
indentation\_consistency & Indentation consistency score \\
space\_pattern\_code & Encoded space/indentation pattern \\
\bottomrule
\end{tabular}
\end{table}

\subsection{Style Encoder}

To encode the extracted style vector into a high-dimensional latent space compatible with the backbone language model, we employ a four-layer multilayer perceptron (MLP) with a residual connection between the second and fourth layers. This design balances representational capacity and training stability.

Formally, given a 34-dimensional feature vector $\mathbf{s}$, the style encoder $f_\theta(\cdot)$ projects it to a 1024-dimensional latent embedding $\mathbf{z}_s$:
\begin{equation}
\mathbf{z}_s = f_\theta(\mathbf{s}) = \text{MLP}_{4\text{-layer}}(\mathbf{s}) + \text{Res}(\mathbf{s})
\end{equation}
where $\text{Res}(\cdot)$ denotes the residual mapping applied across intermediate layers. Specifically, the MLP adopts the following architecture: $34 \rightarrow 128 \rightarrow 512 \rightarrow 768 \rightarrow 1024$ dimensions, totaling approximately 1.65 million parameters. This embedding serves as a conditioning signal injected into the code generator, allowing the model to align output style with the provided reference.

During inference, user-provided style reference code samples—referred to as style exemplars—are analyzed by the style analyzer to produce $\mathbf{s}$, which is then encoded and injected into the LLM. If no prior style reference exists, the model defaults to neutral style generation. This mechanism ensures controllable, user-specific style adaptation during the code generation process \citep{evtikhiev2023out,bui2021self}.

\begin{figure}[ht]
\centering
\includegraphics[width=0.9\textwidth]{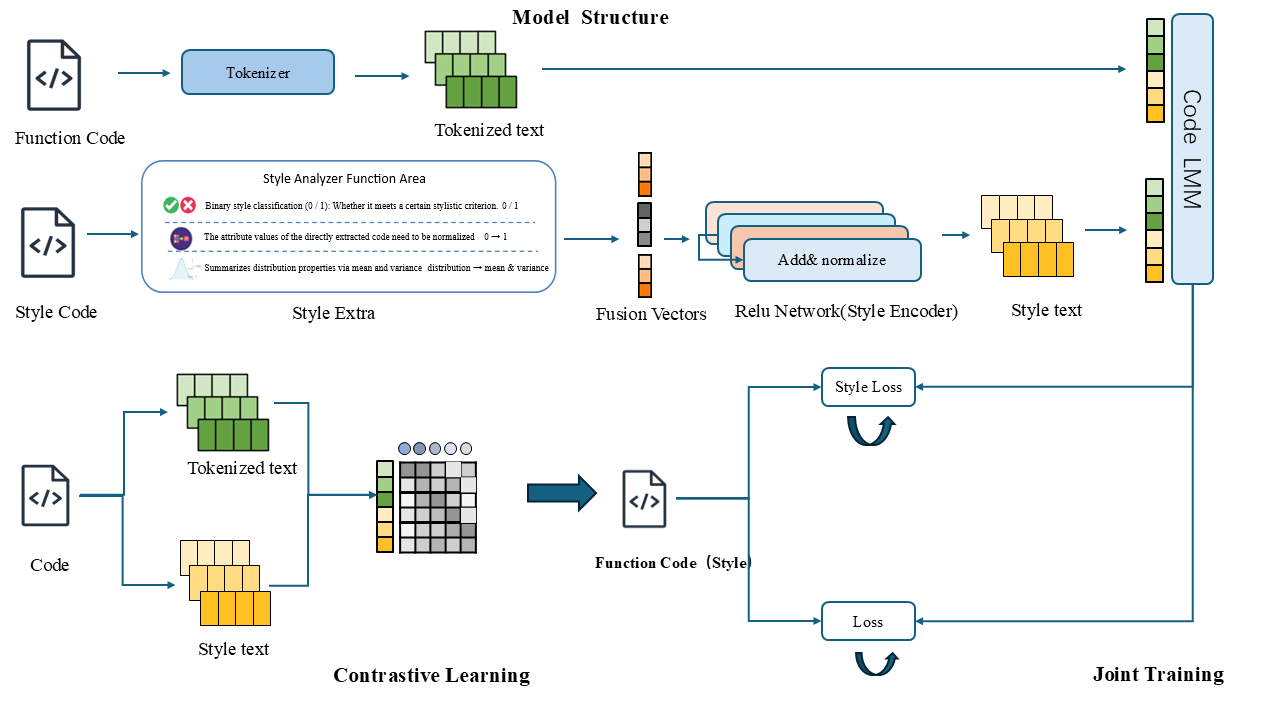}
\caption{Architecture of the Style Encoder. The encoder uses a four-layer MLP with residual connections to project the 34-dimensional style vector to a 1024-dimensional latent embedding compatible with the Flan-T5-Large backbone.}
\label{fig:encoder}
\end{figure}

\section{Training}
\label{sec:training}

We construct our training corpus based on a publicly available dataset \citep{fosso2023python} containing approximately 18K Python code snippets. To ensure style diversity, we use three state-of-the-art code generation models—DeepSeek, Doubao \citep{bytedance2025seed}, and Qwen \citep{yang2025qwen3}—to generate multiple functionally equivalent but stylistically different variants for each original code snippet. The generated code undergoes functional correctness verification, deduplication, and length filtering ($\leq 378$ tokens), ultimately forming a style-diverse dataset containing approximately 337K samples. Each sample pair contains functionally equivalent but stylistically different code pairs, annotated with 34-dimensional style vectors. The dataset is split into training set (approximately 337,212 samples) and validation set (200 samples) in a 99\%/1\% ratio, with a test set of 100-300 samples for final metric evaluation.

\subsection{Contrastive Learning}

In personalized code generation, understanding code style is a challenging task due to its latent and implicit nature. Unlike functional semantics that can be directly inferred from token sequences or abstract syntax trees, code style emerges from a combination of naming patterns, formatting practices, structural choices, and spacing conventions. Therefore, rather than relying on supervised labels or handcrafted style categories, we use a self-supervised contrastive learning strategy that allows the model to automatically discover style patterns from raw code corpora.

The first stage of our framework focuses on learning explicit and discriminative representations of code style through contrastive learning. This component, called the style encoder, is responsible for mapping reference code snippets to a latent style space where stylistically similar samples are embedded close together while dissimilar samples are pushed apart \citep{radford2021learning}.

\subsubsection{Positive and Negative Pair Construction}

To train the style encoder, we construct contrastive pairs from large-scale unlabeled code data. Specifically:

\textbf{Positive pairs} are formed by taking complete code snippets and extracting sub-snippets with identical style characteristics (e.g., function definitions, loop bodies, or comment sections from the same file). These snippets naturally share the same underlying style \citep{nie2024code}.

\textbf{Negative pairs} are formed by pairing code snippets with style snippets sampled from different files or authors to ensure style dissimilarity.

\subsubsection{Loss Function}

We adopt the InfoNCE loss function to train the encoder, defined as:
\begin{equation}
\mathcal{L}_{\text{contrastive}} = -\log\frac{\exp(\text{sim}(\mathbf{z}, \mathbf{z}^+) / \tau)}{\sum_{j} \exp(\text{sim}(\mathbf{z}, \mathbf{z}_j) / \tau)}
\end{equation}
where $\mathbf{z}$ and $\mathbf{z}^+$ are embeddings of the anchor code and its positive style snippet, $\mathbf{z}_j$ are negative samples within the same batch, $\text{sim}(\cdot)$ is the cosine similarity function, and $\tau$ is the temperature hyperparameter.

The style encoder is implemented as a lightweight BiGRU network followed by a linear projection layer. During training, it learns to cluster code snippets with similar styles in the embedding space.

\subsubsection{Training Details}

The contrastive learning process is conducted independently of the language model. The language model is frozen at this stage to ensure that the learned style representations are generic and not entangled with token-level generation patterns. We train the style encoder for 30 epochs using a batch size of 16, temperature $\tau = 0.07$, and embedding dimension of 1024. At this stage, the Flan-T5 model is frozen, with only the style encoder (approximately 1.65 million parameters) being trained. After training, the encoder is fixed and used for style-guided generation in the second stage.

\subsection{Joint Fine-tuning with Style Supervision}

In the second stage, we train a decoder-only generative model called StyleControlledGenerator, which takes as input a source code snippet ($\text{code}_1$) and a target style vector extracted from a reference. We employ a Distributed Data Parallel (DDP) training strategy with an effective batch size of 1, training for 20 epochs. The model is supervised using both semantic loss and style loss.

\textbf{Semantic Loss.} The primary objective is to generate target code ($\text{code}_2$) that matches the ground truth tokens. We apply standard token-level cross-entropy loss:
\begin{equation}
\mathcal{L}_{\text{CE}} = \sum_{t} -\log P(y^{(t)} | y^{(<t)}, \mathbf{x})
\end{equation}
where $\mathbf{x}$ represents the input code and style vector, and $y^{(t)}$ is the $t$-th token in $\text{code}_2$.

\textbf{Style Loss.} To further ensure that the generated code conforms to the expected style, we use the frozen style encoder to extract a style vector from the generated code. We then compute the mean squared error (MSE) between this predicted style vector and the original target style vector:
\begin{equation}
\mathcal{L}_{\text{style}} = \|\text{style\_encoder}(\hat{\mathbf{y}}) - \mathbf{s}_{\text{target}}\|^2
\end{equation}

\textbf{Joint Objective.} The total loss is a weighted sum of the semantic and style objectives:
\begin{equation}
\mathcal{L}_{\text{total}} = \mathcal{L}_{\text{CE}} + \lambda \cdot \mathcal{L}_{\text{style}}
\end{equation}
where $\lambda$ is a hyperparameter controlling the influence of style supervision.

\section{Experiments}
\label{sec:experiments}

We validate the overall performance of our model using 7 model experimental groups and 4 evaluation metrics.

\subsection{Evaluation Metrics}

To comprehensively evaluate the quality and style fidelity of generated code, we employ both standard benchmark metrics and custom-designed metrics:

\textbf{BLEU-4 (from MPCoder).} This metric evaluates n-gram overlap between generated code and reference code, focusing on token-level syntactic similarity. Higher BLEU-4 scores indicate closer lexical matching \citep{evtikhiev2023out}.

\textbf{ROUGE (from MPCoder).} We adopt three ROUGE metric variants to capture different aspects of similarity \citep{sanchan2024comparative}: ROUGE-1 measures unigram overlap, reflecting basic lexical similarity; ROUGE-2 measures bigram overlap, capturing local phrase-level consistency; ROUGE-L evaluates the Longest Common Subsequence (LCS) between generated and reference code, serving as a proxy for structural alignment.

\textbf{CSS (Code Style Similarity).} A task-specific metric designed to quantify style alignment between generated and reference code. It primarily evaluates 14 features of the naming style dimension, including naming conventions (such as snake\_case, camelCase, etc.), case style distribution, and character usage patterns among other surface-level style attributes. CSS focuses on the most intuitively visible style differences, excluding deep structural features.

\textbf{AI Evaluation.} An automated scoring mechanism driven by a language-model-based evaluator. This metric assesses both semantic correctness and style fidelity of generated code, simulating human-like judgment processes.

\subsection{Quantitative Evaluation Results}

To evaluate the performance of our model Style2Code, we compare it with the baseline (Flan-T5) and state-of-the-art models (e.g., DeepSeek, Qwen) using four key evaluation metrics: BLEU-4, AI judgment, ROUGE, and CSS similarity.

\textbf{BLEU-4 Scores.} Our model achieves a BLEU-4 score of 0.317, representing a significant 1450\% improvement over the baseline Flan-T5 (0.020) and an 8.8\% gain over the best-performing DeepSeek (code2$\rightarrow$style1, score of 0.285). It also outperforms the SOTA model MPCoder.

\begin{figure}[ht]
\centering
\includegraphics[width=0.8\textwidth]{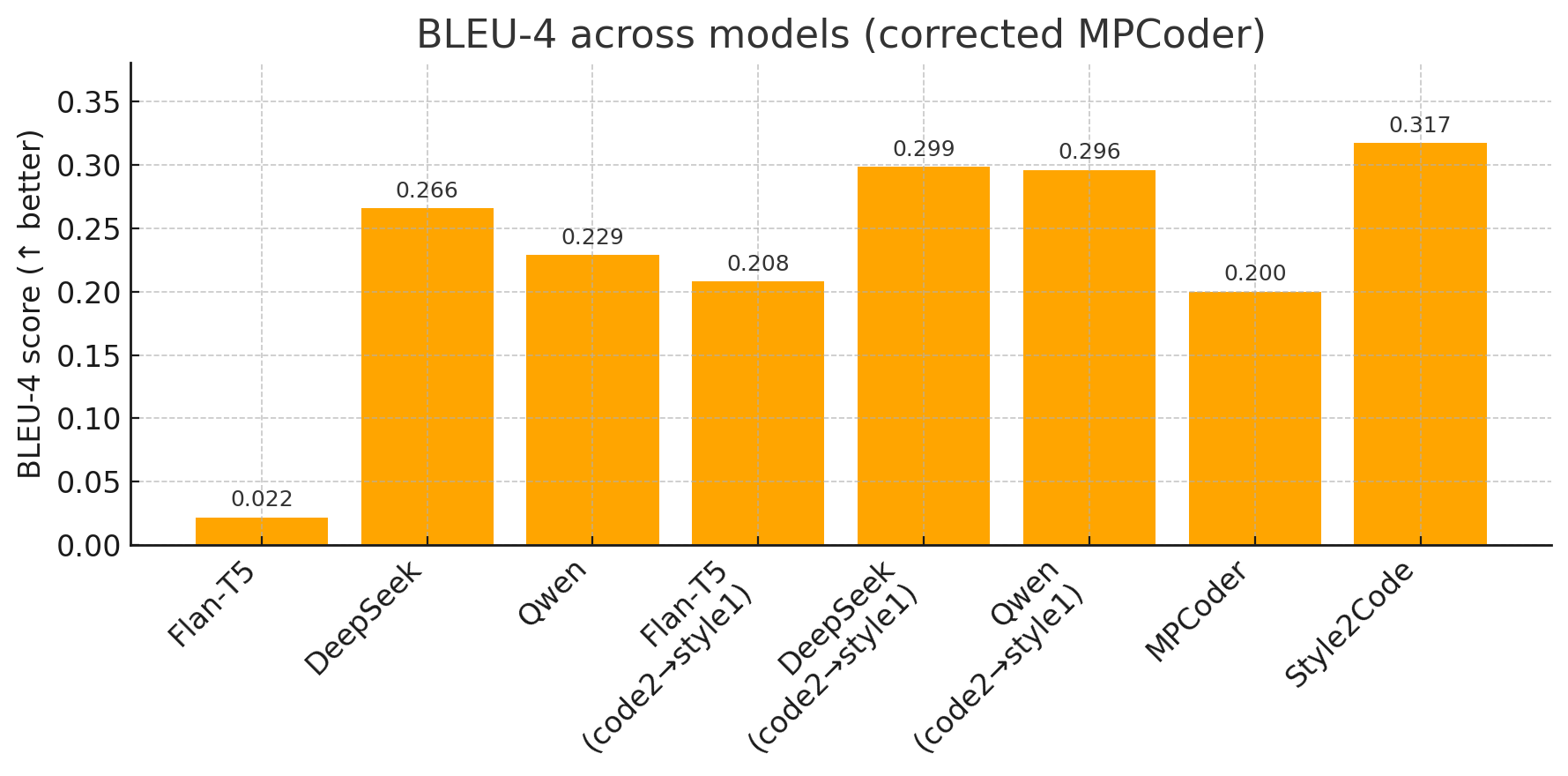}
\caption{BLEU-4 performance comparison across different models. Style2Code achieves a score of 0.317, significantly outperforming the baseline Flan-T5 (0.020, +1450\%) and the best competitor DeepSeek (0.285, +8.8\%).}
\label{fig:bleu}
\end{figure}

\textbf{AI-based Style Judgment.} In the evaluation of style consistency by an AI-based style classifier, Style2Code obtains a score of 0.115, significantly surpassing Flan-T5 (close to 0) and ranking third overall.

\textbf{ROUGE F1 Scores.} Style2Code achieves competitive ROUGE scores across all sub-metrics. Compared to the second-best model MPCoder, Style2Code achieves higher scores on all three ROUGE metrics: ROUGE-1: Style2Code reaches 0.448, exceeding MPCoder (0.371) by +20.8\%; ROUGE-2: Style2Code scores 0.167, improving over MPCoder (0.143) by +16.8\%; ROUGE-L: Style2Code reaches 0.401, slightly higher than MPCoder (0.388) by +3.4\%. Style2Code ranks first among all evaluated models on all ROUGE sub-metrics, demonstrating its strong advantages in style alignment and content similarity.

\begin{figure}[ht]
\centering
\includegraphics[width=0.8\textwidth]{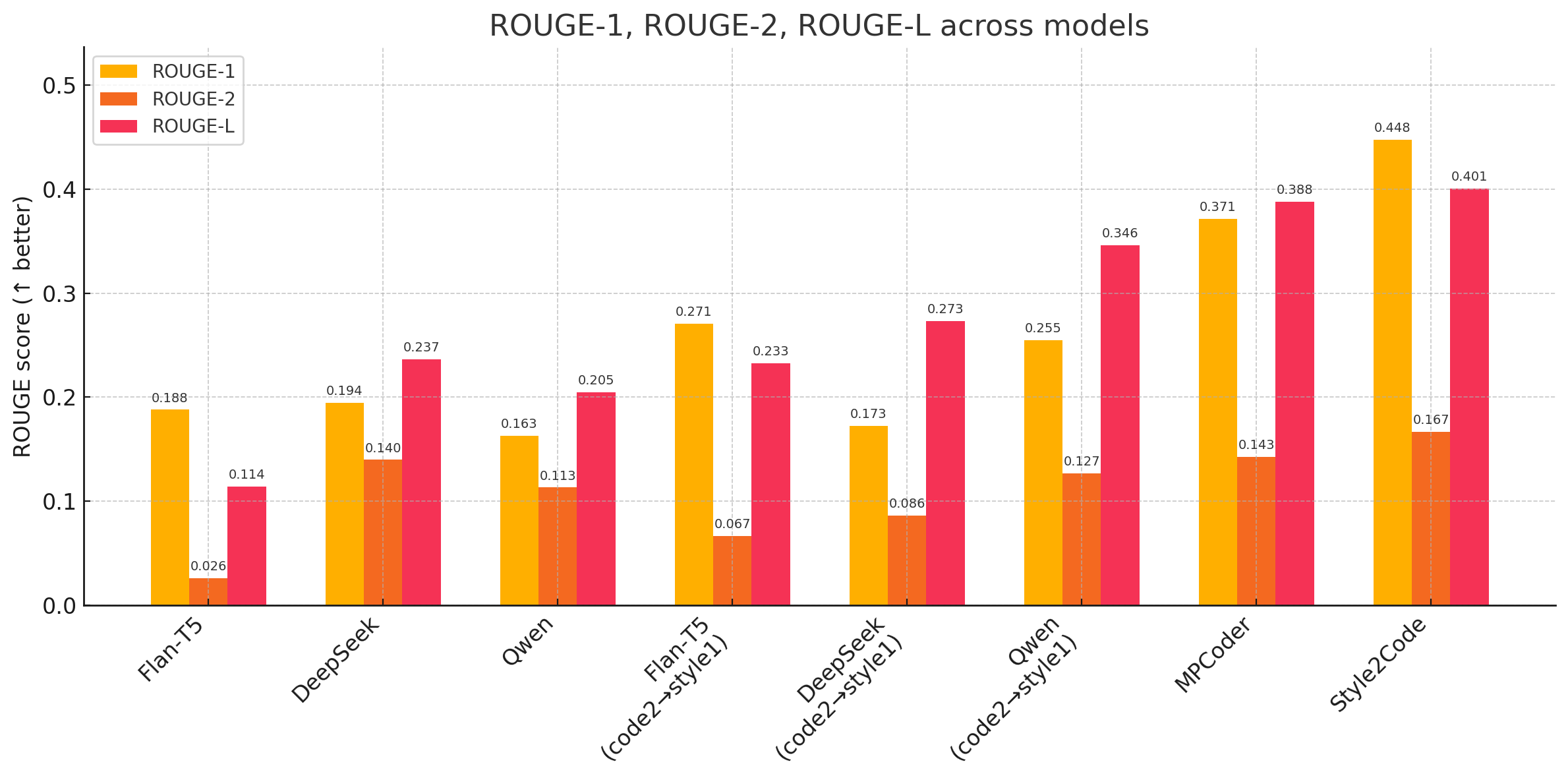}
\caption{ROUGE performance comparison (ROUGE-1, ROUGE-2, ROUGE-L). Style2Code achieves the best performance across all ROUGE metrics: ROUGE-1 (0.448), ROUGE-2 (0.167), and ROUGE-L (0.401), outperforming MPCoder and other baselines.}
\label{fig:rouge}
\end{figure}

\textbf{CSS (Code Style Similarity).} To evaluate style alignment, we employ the custom metric CSS. Style2Code achieves a CSS score of 0.196, ranking third among all evaluated models. The best-performing model Qwen (code2$\rightarrow$style1) reaches 0.222, followed closely by MPCoder (0.214), which serves as a strong SOTA reference in style-conditioned code generation. Compared to MPCoder, Style2Code shows an 8.4\% relative gap but achieves comparable performance without explicit multi-task pretraining. This highlights Style2Code's competitive style control in a more lightweight setting. In contrast, Style2Code exceeds the baseline model Flan-T5 (0.100) by a substantial +96\%, clearly demonstrating the advantages of incorporating style-specific conditioning. The results confirm Style2Code's effectiveness in enforcing style fidelity while maintaining cross-example generalization.

\begin{figure}[ht]
\centering
\includegraphics[width=0.8\textwidth]{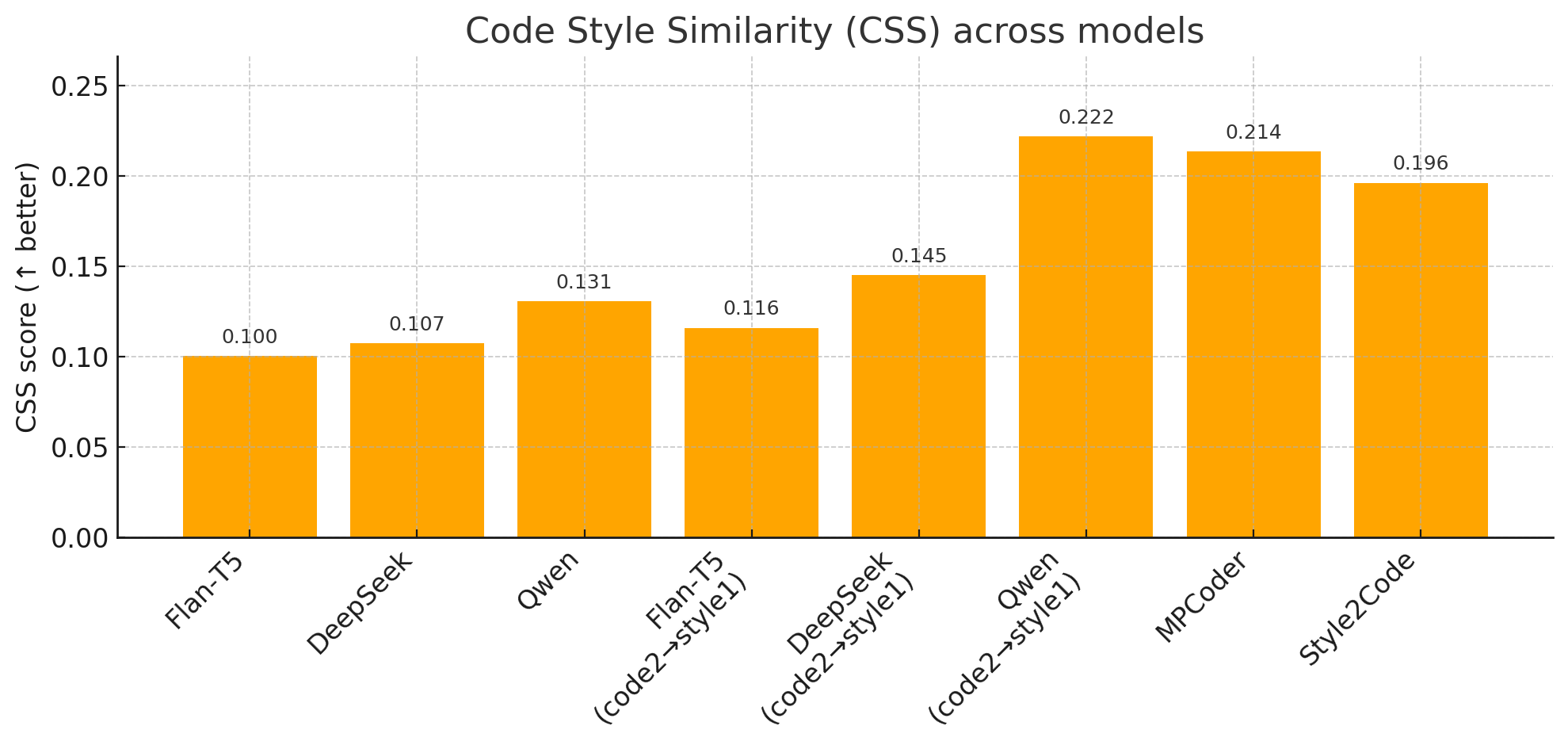}
\caption{CSS (Code Style Similarity) comparison. Style2Code achieves a CSS score of 0.196, ranking third. While Qwen (0.222) and MPCoder (0.214) perform slightly better, Style2Code shows a +96\% improvement over the baseline Flan-T5 (0.100).}
\label{fig:css}
\end{figure}

\begin{figure}[ht]
\centering
\includegraphics[width=0.8\textwidth]{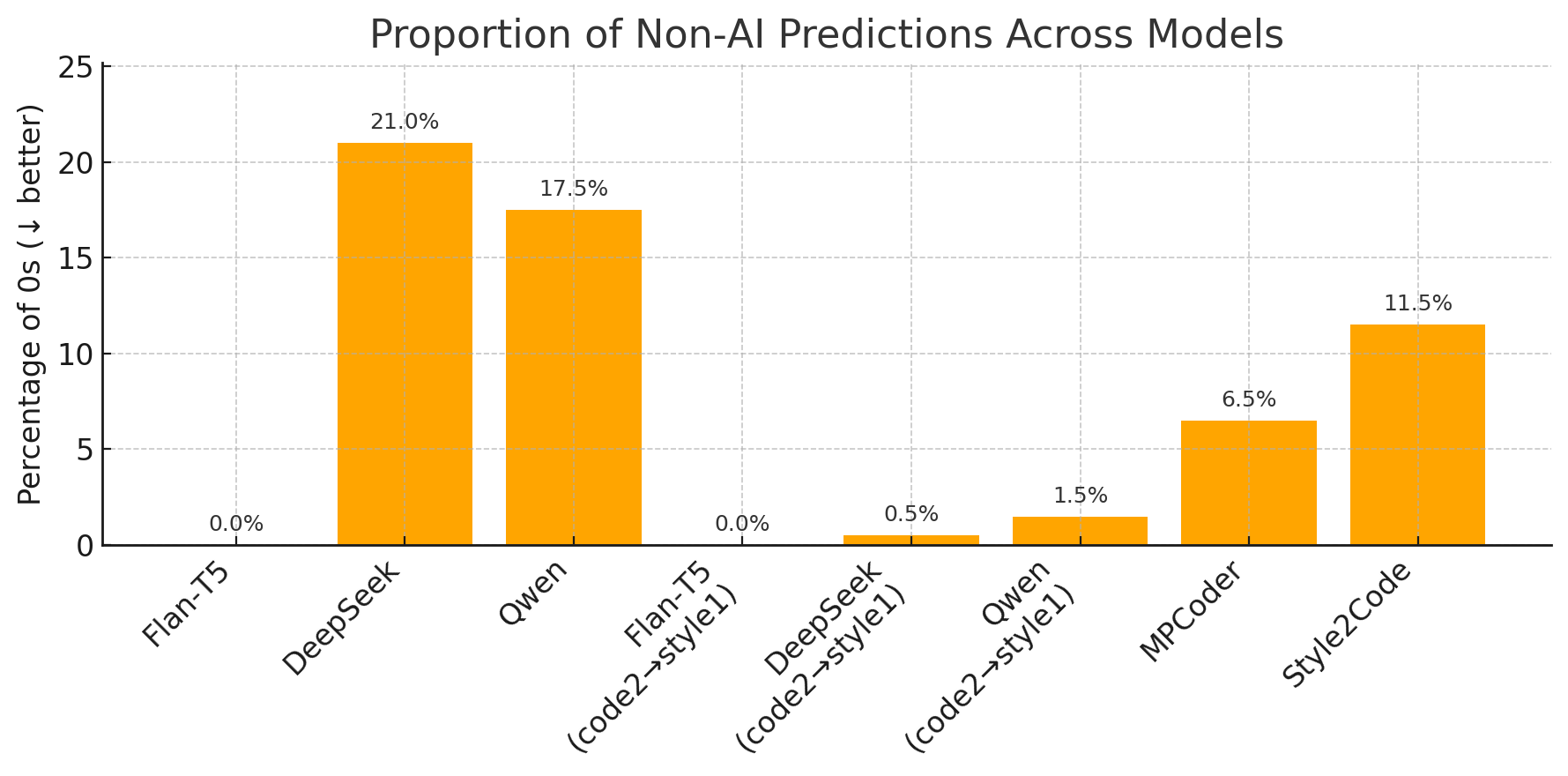}
\caption{AI-based style evaluation scores. Style2Code obtains a score of 0.115, significantly surpassing the baseline Flan-T5 (close to 0) and ranking competitively among state-of-the-art models.}
\label{fig:ai_eval}
\end{figure}

\section{Conclusion}
\label{sec:conclusion}

In this work, we propose a novel style-guided code generation framework that enables generation of source code conforming to specified target styles. This framework addresses the growing need for personalized and style-consistent code in real-world development environments.

Our key innovation lies in using contrastive learning to extract explicit style features from code. Experimental results demonstrate that this approach is effective in capturing fine-grained style characteristics and enhancing the controllability of code generation.

Through extensive evaluation, we show that our model outperforms baselines on multiple metrics, particularly in text-based and fine-grained style matching. Nevertheless, our model still slightly lags behind state-of-the-art (SOTA) models on structure-aware metrics due to the lack of targeted learning strategies for implicit structural features.

An important advantage of our approach is its complementary effect: while large language models typically excel at generating structurally sound code through implicit prompting, our model complements this by enhancing fine-grained style control through explicit guidance, forming a more complete solution for code style alignment.

Furthermore, our implementation (based on Flan-T5-Large, approximately 850 million parameters) demonstrates strong performance, emphasizing the scalability and practical effectiveness of the proposed method in large model settings. By treating code style as an explicit modality, our framework offers a promising direction in the field of dual-modal learning, where conditioning based on fine-grained, learnable modalities remains a key challenge. This formulation not only improves the controllability of code generation but also introduces a generalizable approach for modality-aware code synthesis.

\subsection{Limitations}

Despite significant progress in style control, Style2Code has some limitations. First, our 34-dimensional explicit style vector may not fully capture certain deep, implicit style features such as algorithmic design preferences or high-level architectural choices. Second, the current data construction pipeline relies on existing code generation models (DeepSeek, Doubao, Qwen), which may introduce inherent biases from these models. Finally, our approach has currently only been evaluated on Python code, and its generalization ability to other programming languages requires further verification.

\subsection{Future Work}

We plan to extend this work in several directions: (1) explore richer style representation methods that combine implicit learning and explicit feature extraction; (2) extend the framework to multiple programming languages and study cross-language style transfer; (3) introduce user feedback mechanisms to allow interactive style adjustment and optimization; (4) investigate how to combine style control with code quality assurance (such as readability and maintainability).

Finally, we note that the field of code style transfer remains underexplored. Existing methods are mostly limited to agent-based multi-turn dialogue or rely on multi-modal strategies with identical content. Our unified single-model architecture represents a new theoretical direction for this task, helping to advance the methodology and applications in this field.

\section*{Acknowledgements}
We thank the anonymous reviewers for their valuable feedback.

\section*{Declarations}

\noindent\textbf{Funding:} Not applicable.

\noindent\textbf{Conflict of interest:} The authors declare no competing interests.

\noindent\textbf{Ethics approval:} Not applicable.

\noindent\textbf{Data availability:} Source code and dataset are available at \url{https://github.com/zh19980811/Style2Code}

\noindent\textbf{Code availability:} Available at the same repository.

\bibliographystyle{plainnat}
\bibliography{references}

\end{document}